\documentclass[11pt,a4paper]{article}
\usepackage[hyperref]{emnlp-ijcnlp-2019}
\usepackage{times}
\usepackage{latexsym}
\usepackage{booktabs}
\usepackage{amsmath}
\usepackage{amssymb}
\usepackage{caption}
\usepackage{tikz}
\usepackage{subcaption}
\usepackage{pgfplots}
\usetikzlibrary{decorations.text}
\usetikzlibrary{positioning}
\usepackage{mathtools}
\DeclarePairedDelimiter{\ceil}{\lceil}{\rceil}
\usepackage[normalem]{ulem}
\usepackage{enumitem}
\usepackage{url}
\usepackage{tablefootnote}
\usepackage{soul}

\aclfinalcopy % Uncomment this line for the final submission

\title{Non-Autoregressive Machine Translation with Latent Alignments}

\author{Chitwan Saharia\Thanks{\;Equal contribution.} \,\Thanks{\;Work done as part of the Google AI Residency.} \\
  Google Research, Brain Team \\
  {\tt sahariac@google.com} \\\And
  William Chan\footnotemark[1] \\
  Google Research, Brain Team \\
  {\tt williamchan@google.com} \\\AND
  Saurabh Saxena \\
  Google Research, Brain Team \\
  {\tt srbs@google.com} \\\And
  Mohammad Norouzi \\
  Google Research, Brain Team \\
  {\tt mnorouzi@google.com}}
\date{}

\begin{document}
\maketitle
\begin{abstract}
This paper presents two strong methods, CTC and Imputer, for non-autoregressive
machine translation that model latent alignments with dynamic programming. We revisit CTC for machine translation and demonstrate that a simple CTC model can achieve state-of-the-art for single-step non-autoregressive machine translation, 
contrary to what prior work indicates. 
In addition, we adapt the Imputer model for non-autoregressive machine translation and demonstrate that Imputer with just 4 generation steps can match the performance of an autoregressive Transformer baseline.
Our latent alignment models are simpler than many existing non-autoregressive translation baselines; for example, we do not require target length prediction or re-scoring with an autoregressive model.
On the competitive WMT'14 En$\rightarrow$De task, our CTC model achieves 25.7 BLEU with a single generation step, while Imputer achieves 27.5 BLEU with 2 generation steps, and 28.0 BLEU with 4 generation steps. This compares favourably to the autoregressive Transformer baseline at 27.8 BLEU.
\end{abstract}

\section{Introduction}

Non-autoregressive neural machine translation \citep{gu-iclr-2018} 
aims to enable the parallel generation of output tokens
without sacrificing translation quality.
There has been a surge of recent interest in this family of
efficient decoding models,
resulting in the development of
iterative refinement~\citep{lee-emnlp-2018},
CTC models \citep{libovicky-emnlp-2018},
insertion-based methods \citep{stern-icml-2019, chan-arxiv-2019}, edit-based methods \cite{gu-neurips-2019,ruis-wngt-2019}, masked language models \citep{ghazvininejad-emnlp-2019, ghazvininejad-arxiv-2020}, and normalizing flow models~\citep{ma2019flowseq}.
Some of these methods generate the output tokens in a constant number
of steps~\citep{gu-iclr-2018, libovicky-emnlp-2018,
  lee-emnlp-2018, ghazvininejad-emnlp-2019, ghazvininejad-arxiv-2020}, while others
require a logarithmic number of generation steps~\citep{stern-icml-2019,chan-arxiv-2019,chan-neurips-2019,li-wngt-2019}.

Recent progress has decreased the gap between autoregressive
and non-autoregressive models' translation scores.
%Despite recent advances, 
However, non-autoregressive models often suffer from two main limitations:
\begin{enumerate}[topsep=0pt, partopsep=0pt, leftmargin=12pt, parsep=0pt, itemsep=2pt]
\item First, most non-autoregressive models assume that the output tokens are conditionally independent given the input.
This leads to the weakness of such models in generating multi-modal outputs~\citep{gu-iclr-2018}, and materializes in the form of \textit{token repetitions} in the decoded outputs.
Addressing this limitation generally involves stochastic search algorithms like noisy parallel decoding \citep{gu-iclr-2018}, iterative decoding \citep{ghazvininejad-emnlp-2019,ghazvininejad-arxiv-2020}, or simple but less effective heuristic methods such as collapsing repetitions \citep{lee-emnlp-2018}.
\item
The second limitation of many prior non-autoregressive models is the requirement of \textit{output length prediction} as a pre-process.
Autoregressive models have the ability to dynamically adjust the output sequence length by emitting an \texttt{<END>}
token at any generation step to stop.
Many non-autoregressive models often require a fixed length decoder. Thus they train a separate target length prediction module,
and at inference time, first predict and condition on the target length, and then generate the output tokens \citep{gu-iclr-2018}.
Since the model needs to commit to a fixed predicted length, which cannot be changed dynamically,
it is often required to use multiple length candidates and re-score them to produce the final translation~\cite{ghazvininejad-emnlp-2019,ghazvininejad-arxiv-2020}.
\end{enumerate}

This paper addresses the limitations of existing non-autoregressive machine translation models by using
\textit{latent alignment models}. Latent alignment models utilize a sequence of discrete latent alignment variables
to monotonically align the non-autoregressive predictions of the model and output tokens.
Such models use dynamic programming to marginalize out the alignment variables during training.
This paper studies two instances of latent alignment models including Connectionist Temporal Classification 
(CTC)~\citep{graves-icml-2006, graves-icassp-2013, graves-icml-2014} and Imputer \citep{chan2020imputer}.
\citet{libovicky-emnlp-2018} have previously applied CTC to non-autoregressive machine translation.
However, we report a significant improvement over the work of \citet{libovicky-emnlp-2018} and demonstrate that CTC can achieve the state-of-the-art in single-step non-autoregressive machine translation.
We attribute this performance difference primarily to our use of distillation during training, similar to \citet{gu-iclr-2018}.
We adapt latent alignment models to machine translation and demonstrate their effectiveness
on non-autoregressive machine translation, advancing state-of-the-art on WMT'14 En$\leftrightarrow$De
and WMT'16 En$\leftrightarrow$Ro. 

The main contributions of this paper include:
\begin{enumerate}[topsep=0pt, partopsep=0pt, leftmargin=12pt, parsep=0pt, itemsep=2pt]
    \item We adapt latent alignment models to non-autoregressive machine translation.
    \item We achieve a new state-of-the-art of 25.8 BLEU on WMT'14 En$\rightarrow$De for single step non-autoregressive machine translation.
    \item We achieve 27.5 BLEU with 2 step generation, 28.0 BLEU with 4 step generation, and 28.2 BLEU with 8 step generation for WMT'14 En$\rightarrow$De, setting a new state-of-the-art for non-autoregressive machine translation with a constant number of generation steps.
\end{enumerate}

\begin{figure*}[t]
\small
\begin{flushleft}
\textbf{Source:} Ein weiterer, besonders wichtiger Faktor sei die Vernetzung von Hochschulen und Unternehmen. \\[.5em]

\textbf{Imputer Decoding:}
\end{flushleft}
\vspace{-.8em}
\begin{center}
\resizebox{\textwidth}{!}{%
\input{paper_decode_example.tex}
}\\[.5em]
\begin{flushleft}
\textbf{Output:} Another particularly important factor is the networking of universities and businesses.
\end{flushleft}
\caption{\small Example of top-\textit{k} decoding using Imputer with 8 decoding steps.  For a sentence of length $N$, the model imputes $\ceil{\frac{N}{8}}$ tokens at every decoding step. In each row, blue underlined tokens are the ones being imputed. Tokens that are not generated yet are colored gray. Note that ``\_'' represents the special blank token that is removed for generating the final target sentence. }

\label{fig:decoding}
\end{center}
\end{figure*}

\section{Latent Alignment Models}
\label{sec2}

We begin by describing the notion of \textit{alignment}, which in the context of this paper is defined as in the CTC literature  \citep{graves-icml-2006,graves-icassp-2013,graves-icml-2014}
and should not be confused with word alignments in machine translation \cite{manning1999foundations,dyer-naacl-2013}.
Alignment is a mapping between a sequence of predicted tokens and a sequence of target tokens. Alignment can be constructed by inserting special ``blank tokens'' into the target sequence to match a pre-specified length. Our alignments have the same length as the source sequences, and collapsing the alignment's blank tokens will recover the target sequence.

Let $x$ denote a source sequence and let $y$ denote a target sequence, where $y_i \in \mathcal{V}$ and $\mathcal{V}$ is the target vocabulary. We make two assumptions: 1) there exists a monotonic mapping between the model's predictions and the target sequence, and 2) the source sequence is at least as long as the target sequence, i.e. $|x| \geq |y|$. 
We define an alignment $a$ between $x$ and $y$ as a discrete sequence in which $a_i \in \mathcal{V} \cup \{``\_"\}$, $|a| = |x|$, and $``\_"$ is a special ``blank" token that is removed to convert $a$ to the target sequence $y$. %This blank token conforms to the definition from prior CTC literature \citep{graves-icml-2006}.
We define a function $\beta(y)$ that returns all possible alignments for a sequence $y$ of a particular length $|x|$. We also define the collapsing function $\beta^{-1}(a)$ such that $\beta^{-1}(a) = y$ if $a \in \beta(y)$.
To avoid token repetitions, it is useful to define the collapsing function $\beta^{-1}(a)$ as first collapsing all consecutive repeated tokens, and then removing all blank tokens.
This formulation follows CTC precisely \cite{graves-icml-2006}.
For instance, given a source sequence $x$ of length 10, and a target sequence $y = (A,A,B,C,D)$, then a possible alignment $a$ is $(\_,A,A,\_,A,B,B,C,\_,D)$. 

The log-likelihood of the target sequence is recovered by marginalizing the latent alignments:
\begin{align}
\log p_\theta(y|x) = \log \sum_{a \in \beta(y)} p_\theta(a|x)
\label{eqn: maginalize_alignments}
\end{align}
The summation in \eqref{eqn: maginalize_alignments} is typically intractable, since there are a combinatorial number of alignments. In the next two sub-sections, we will briefly describe two variants of latent alignment models that leverage dynamic programming to tractably compute the log-likelihood, Connectionist Temporal Classification (CTC) \citep{graves-icml-2006} and Imputer \citep{chan2020imputer}.

\subsection{Connectionist Temporal Classification}
Connectionist Temporal Classification (CTC) \citep{graves-icml-2006,graves-icassp-2013,graves-icml-2014} models the alignment distribution with a strong conditional independence assumption:
\begin{align}
  p_\theta(a|x) = \prod_{i} p(a_i|x;\theta)
\end{align}

Leveraging this strong conditional independence assumption enables CTC to use an efficient dynamic programming algorithm to exactly marginalize out the latent alignments:
\begin{align}
\log p_\theta(y|x) = \log \sum_{a \in \beta(y)} \prod_{i} p(a_i|x;\theta)
\end{align}
This allows us to compute the log-likelihood and its gradient tractably.
We refer the reader to \citet{graves-icml-2006} for the exact details of the dynamic programming algorithm.
During inference, CTC generates the alignment distribution in parallel with a single generation step; the output sequence can then be recovered by greedy decoding or beam search \citep{graves-icml-2006}. We use greedy decoding in all our experiments.

\begin{figure*}[t]
    \centering
    \begin{subfigure}[t]{0.33\textwidth}
    \centering
    \scalebox{.8}{\begin{tikzpicture}[node distance=1cm, scale=1.0, every node/.style={transform shape}]
\tikzstyle{layer} = [rectangle, thick, rounded corners, minimum width=4cm, minimum height=0.5cm, align=center, draw=black]
\tikzstyle{symbol} = [align=center, minimum height=0.6cm]
\tikzstyle{arrow} = [thick,->]

\node (posterior_) [symbol,align=center] at (10.5, 1.25) {Alignment Probability \\ $p_\theta(a | x)$};
\node (d3_) [layer] at (10.5,0) {Softmax};
\node (d2_) [layer, below of=d3_] {Self-Attention};
\node (d1_) [layer, below of=d2_] {Upsample};
\node (d0_) [layer, below of=d1_] {Embedding};
\node (source_) [symbol, below of=d0_] at (10.5, -3.2) {$x$: Source Sequence};

\draw[arrow] (source_) -- (d0_);
\draw[arrow] (d0_) -- (d1_);
\draw[arrow] (d1_) -- (d2_);
\draw[arrow] (d2_) -- (d3_);
\draw[arrow] (d3_) -- (posterior_);

\end{tikzpicture}}
    \caption{CTC}
    \label{subfig:ctc}
    \end{subfigure}%
    \begin{subfigure}[t]{0.66\textwidth}
    \centering
    \scalebox{.8}{\begin{tikzpicture}[node distance=1cm, scale=1.0, every node/.style={transform shape}]
\tikzstyle{layer} = [rectangle, thick, rounded corners, minimum width=4cm, minimum height=0.5cm, align=center, draw=black]
\tikzstyle{symbol} = [align=center, minimum height=0.6cm]
\tikzstyle{arrow} = [thick,->]

\node (d3) [layer] at (2.5,0) {Softmax};
\node (d2) [layer, below of=d3] {Self-Attention};
\node (d1) [symbol, below of=d2] {$+$};
\node (d0) [layer] at (0, -3.0) {Upsample};
\node (d)  [layer] at (0, -4.0) {Embedding};
\node (e0) [layer] at (5,-3.5) {Embedding};
\node (source) [symbol] at (0, -5.0) {$x$: Source Sequence};
\node (hyp) [symbol,align=center] at (5, -5.0) {$\tilde a$: Prior Alignment \\ (contains masked out tokens)};
\node (posterior) [symbol,align=center] at (2.5, 1.25) {Alignment Probability \\ $p_\theta(a | x, \tilde a)$};

\draw[arrow] (source) -- (d);
\draw[arrow] (d) -- (d0);
\draw[arrow] (d0.north) -- (0, -2.0) -- (d1.west);
\draw[arrow] (d1) -- (d2);
\draw[arrow] (d2) -- (d3);
\draw[arrow] (hyp.north) -- (e0.south);
\draw[arrow] (e0.north) -- (5, -2.0) -- (d1.east);
\draw[arrow] (d3.north) -- (posterior.south);

\end{tikzpicture}}
    \caption{Imputer}
    \label{subfig:imputer}
    \end{subfigure}
    \caption{Visualization of the CTC (\subref{subfig:ctc}) and Imputer (\subref{subfig:imputer}) architecture for non-autoregressive machine translation.}
    \label{fig:architecture}
\end{figure*}
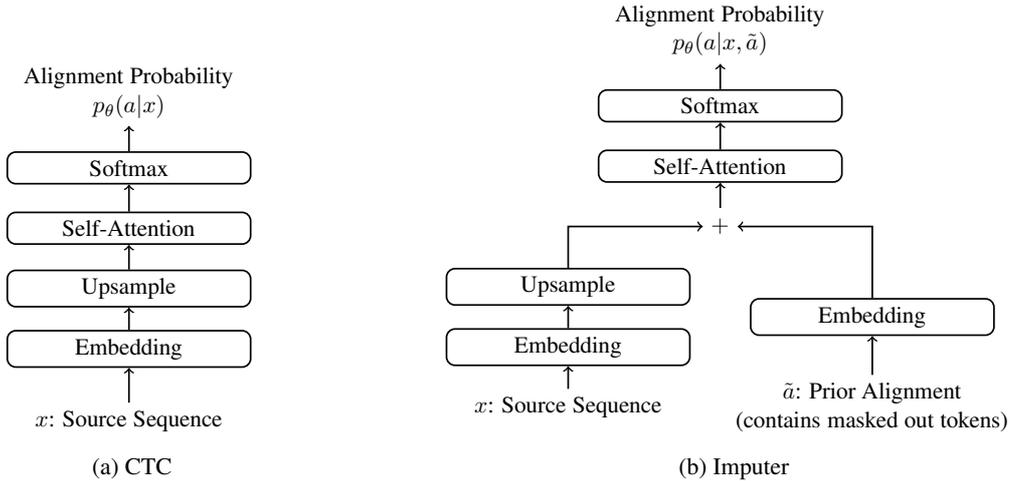
\subsection{Imputer}
The CTC model makes strong conditional independence assumption between alignment token predictions. This assumption licenses CTC to generate the entire alignment in parallel, with a single generation step independent of the number of source or target tokens. However, the strong conditional independence assumption limits its capacity to model complex multi-modal distributions. On the other hand, autoregressive models are capable of modelling such complex multi-modalities with the chain rule factorization, but requires $n$ decoding steps to generate $n$ tokens during inference. Imputer \cite{chan2020imputer} aims to address these limitations.

Imputer is an iterative generative model needing only a constant number of generation steps for inference. It makes conditional independence assumptions within a generation step to achieve parallel generation, and models conditional dependencies across generation steps. This approach has been applied successfully in speech recognition \cite{chan2020imputer}, matching autoregressive models with only a constant number of generation steps. Imputer models the distribution of alignments $p_\theta(a|x)$ as:
\begin{align}
    p_\theta(a|x) = \sum_{\tilde a \in \gamma(a)} p_\theta(a|\tilde a, x) p(\tilde a | x)
    \label{eqn: marginalize_masks}
\end{align}

where $\tilde a$ is a (partially masked out) alignment, and $\gamma(a)$ is the set of all possible masking permutations of $a$. \eqref{eqn: marginalize_masks} marginalizes over all possible alignments between the input and output sequences, and all possible generation orders. Imputer models the next alignment $a$ conditioned on the previous alignment $\tilde a$:
\begin{align}
    p_\theta(a | \tilde a, x) = \prod_i p(a_i | \tilde a, x; \theta)
\end{align}
The key insight to Imputer is that we can construct a log-likelihood lower-bound:
\begin{equation}
\begin{aligned}
    & \log p_\theta(y|x) \\
    & \geq \mathbb{E}_{a \sim \beta(y)} \left[ \mathbb{E}_{\tilde a \sim \gamma(a)} \left[ \log \sum_{a' \in \beta'(\tilde a, a)} p_\theta(a'|\tilde a, x) \right] \right]
    \label{eqn:imputer_dp}
\end{aligned}
\end{equation}
where $a' \sim \beta'(\tilde a, a)$ captures all possible alignments $a'$ consistent with $(\tilde a, a)$ \citep{chan2020imputer}. This equation can be solved efficiently via dynamic programming \citep{chan2020imputer}.
This formulation licenses Imputer with an iterative generation process. Tokens are generated independently (and in parallel) within a generation step but are conditioned on the partially predicted alignment $\tilde a$ of the last iteration (unlike CTC). In practice, Imputer uses a constant number of decoding iterations independent of the sequence length \citep{chan2020imputer}.

% \subsection{Advantages}

Both CTC and Imputer have seen much success in tasks like speech recognition \cite{graves-icml-2014,chan2020imputer}. However, to the best of our knowledge, these latent alignment models have not been widely applied to machine translation, with the exception of \citet{libovicky-emnlp-2018}. These latent alignment models hold two key advantages over prior non-autoregressive machine translation work \citep{gu-iclr-2018,ghazvininejad-emnlp-2019}, namely: the \textit{token repetition problem} and the \textit{target length prediction problem}. We will discuss them in detail in Section 3.

\section{Latent Alignment Models for Machine Translation}
\label{sec:adapt_la}

In this section, we will discuss how latent alignment models can be adapted to machine translation, and then describe key advantages offered by these models. Section~\ref{sec2} identified two assumptions made by latent alignment models: 1) there exists a monotonic mapping between the model alignment predictions and the target sequence, and 2) the length of the target sequence is less than or equal to the length of source sequence, i.e. $|y| \leq |x|$.  We will now address these issues to adapt our models for machine translation.

\textbf{Monotonic Assumption.} A monotonic structure between model alignment predictions and the target sequence is desired for the dynamic programming algorithm to marginalize out the latent alignments in Equation \eqref{eqn: maginalize_alignments}. Unlike tasks such as speech recognition, a monotonic relationship between the model alignment predictions and the target sequence may not exist in machine translation. For instance, speech-to-text is inherently local, whereas there is typically some global word reordering in machine translation. We hypothesize that if we use a powerful deep neural network like the Transformer \cite{vaswani-nips-2017}, the Transformer will have sufficient computational capacity to learn to reorder the contextual embeddings such that it is approximately monotonic with the target sequence. \citet{libovicky-emnlp-2018} also made a similar assumption.

\textbf{Length Assumption.} By construction, our alignments are the same length as the source sequence, and consequently, we can not generate a target sequence longer than the source sequence. This is not a problem for speech recognition, since the source sequence is generally much longer than the target sequence; however, this is prohibitively restrictive for machine translation. This issue was also discussed in \citet{libovicky-emnlp-2018}, and they proposed a simple solution of up-sampling the source sequence to $s$ times the original length. Choosing a sufficient canvas scale of $s$, we can ensure the alignment is long enough to model the target sequence across our training distribution. We use a very similar up-sampling method applied to the embedding matrix of the source sequence. Given a source sequence embedding $x \in \mathbb{R}^{|x| \times d}$ with $d$-dimension and length $|x|$, we simply up-sample $x' \in \mathbb{R}^{s\cdot|x| \times d}$ via an affine transformation.

\subsection{Model Architecture}
Our neural architecture is simply a stack of self-attention layers \citep{vaswani-nips-2017}.
The source sequence is upsampled (to handle longer target sequences as described above).
In the Imputer architecture, the input to our self-attention stack is simply the superpositioning of the upsampled source and the previous alignment. Our work differs from the prior method, 1) our unified architecture does not have separate encoder decoders which require cross-attention mechanisms, 2) our architecture is bidirectional, and does not rely on causality masks. Figure \ref{fig:architecture} visualizes our architecture.

\begin{table*}[t]
\centering
\caption{BLEU comparison for various single step generation models. Our simple CTC model is able to outperform all prior single step generation models. $^\dagger$The main difference between our CTC model and prior work \citep{libovicky-emnlp-2018} is that we use data distillation. }
\label{table:1_step_decoding}
%\scalebox{0.72}{
\small
\begin{tabular}{lcccccc}
\toprule
& &   \multicolumn{2}{c}{\textbf{WMT'14}} & \multicolumn{2}{c}{\textbf{WMT'16}} \\
\bfseries Method & \bfseries Iterations & \bfseries En$\rightarrow$De & \bfseries De$\rightarrow$En & \bfseries En$\rightarrow$Ro & \bfseries Ro$\rightarrow$En \\ %& \bfseries En$\rightarrow$Zh \\
\midrule
\textit{Non-Autoregressive} \\
\quad Iterative Refinement \citep{lee-emnlp-2018} & 1 & 13.9 & 16.7 & 24.5 & 25.7 \\
\quad NAT with Fertility \citep{gu-iclr-2018} & 1 & 17.7 & 21.5 & 27.3 & 29.1 \\
\quad CTC$^\dagger$ \citep{libovicky-emnlp-2018} & 1 & 17.7 & 19.8 & 19.9 & 24.7 \\
\quad Mask-Predict \citep{ghazvininejad-emnlp-2019} & 1 & 18.0 & 19.3 & 27.3 & 28.2 \\
\quad SMART \citep{ghazvininejad-arxiv-2020} & 1 & 18.6 & 23.8 & - & - \\
\quad Auxiliary Regularization \citep{wang2019non} & 1 &  20.7 & 24.8 & - & -\\
\quad Bag-of-ngrams Loss \citep{shao2019minimizing} & 1 & 20.9 & 24.6 & 28.3 & 29.3 \\
\quad Hint-based Training \citep{li2019hint} & 1 & 21.1 & 25.2 & - & - \\
\quad FlowSeq \citep{ma2019flowseq} & 1 &  21.5 & 26.2 & 29.3 & 30.4 \\
\quad NAT (TCL) \citep{liu2020task} & 1 & 21.9 & 25.6 & - & - \\
\quad Bigram CRF \citep{sun2019fast} & 1 & 23.4 & 27.2 & - & - \\
\quad AXE CMLM \citep{ghazvininejad-axe-cmlm} & 1 & 23.5 & 27.9 & 30.8 & 31.5 \\
\quad NAT (EM + ODD) \citep{sun2020approach} & 1 & 24.5 & 27.9 & - & - \\

\midrule
\textit{Our Work} \\
\quad CTC$^\dagger$   & 1  & 25.7 & 28.1 & 32.2 & 31.6 \\
\quad Imputer   & 1  & \textbf{25.8} & \textbf{28.4} & \textbf{32.3} & \textbf{31.7}  \\
\bottomrule
\end{tabular}
%}
\end{table*}

\subsection{Advantages}

Latent alignment models mitigate two common issues shared by many non-autoregressive machine translation models --  \textit{token repetition} and the requirement for \textit{separate target length prediction}.

\subsubsection{Fewer Token Repetitions}
\label{token_repetitions}
Non-autoregressive sequence models make a conditional independence assumption between token predictions. This licenses them to parallel token generation during inference; however, it makes it difficult to model complex multi-modal distributions. This is especially true for single-step generation models which make strong conditional independence assumptions. During inference, this conditional independent generation often results in the \textit{token repetition problem}, where tokens are erroneously repeated in the output sequence.

This issue has been discussed extensively in prior works \citep{gu-iclr-2018, lee-emnlp-2018, ghazvininejad-emnlp-2019} in the context of machine translation, and has been shown to have a negative impact on performance. To handle these repetitions, \citet{gu-iclr-2018} used Noisy Parallel Decoding, wherein they sample a large number of translation hypotheses and use an autoregressive teacher to re-score them to implicitly penalize translations with more erroneous repetitions. \citet{lee-emnlp-2018} adopted a simple but less effective heuristic of simply removing all consecutive repetitions from the predicted target sequence. \citet{ghazvininejad-emnlp-2019} hypothesized that iterative decoding can help remove repetitions by allowing the model to condition on parts of the input, thus collapsing the multi-modal distribution into a sharper uni-modal distribution. They empirically show that the first few decoding iterations are crucial for removing repetitions resulting in a sharp increase in performance. 

Like other non-autoregressive models, latent alignment models also perform conditionally independent generation, and hence face the issue of token repetitions. Although they differ from the other models in that they do not generate the target sequence directly. Rather, the inference process involves the generation of the target alignment, followed by collapsing the generated alignment into the target sequence using the collapsing function $\beta^{-1}$. Recall by construction, $\beta^{-1}$ collapses repeated tokens \citep{graves-icml-2006}, this inference process enables these models to handle erroneous repetitions implicitly by naturally collapsing them. In particular, for single-step decoding, we show that our CTC based model %implicitly
removes most of the repetitions while collapsing the alignment into target sequence, resulting in a significant improvement in translation quality over prior single step generation models. In addition, we show that Imputer requires just 4 decoding iterations to achieve state-of-the-art translation scores on WMT14 En$\rightarrow$De, in contrast to 10 iterations used by Mask-Predict~\citep{ghazvininejad-emnlp-2019}.  

\subsubsection{No Target Length Prediction Needed}
Many prior non-autoregressive models \citep{gu-iclr-2018,ghazvininejad-emnlp-2019} first predict the target length, then conditioned on the target length predict the target sequence. This is needed because these architectures utilize an encoder-decoder formulation, and the decoder requires a fixed canvas size to work with. The length is fixed, and it cannot be changed dynamically by the model during decoding. Due to this lack of flexibility, during inference, one typically samples multiple length candidates and performs decoding for each length followed by re-ranking them to get a final translation. This not only requires tuning of a new hyperparameter for determining the number of length candidates to sample during inference but also entails a considerable amount of extra inference computation.
 
Our latent alignment models do not require target length prediction, but rather implicitly determine the target sequence length through the alignment. This is possible since the alignment is of the same length as the source sequence, thus eliminating the requirement of predicting target length in advance during inference. The caveat is that we can not generate a target sequence longer than the source sequence, which we address in Section \ref{sec:adapt_la}. \citet{libovicky-emnlp-2018}, which also applied CTC to machine translation, made a similar argument, and we further extend this to Imputer. Our approach simplifies the architecture and decoding process, avoiding a need to build a target length prediction model and searching over it during inference.

\section{Related Work}
There has been significant prior work on non-autoregressive iterative methods for machine translation \citep{gu-iclr-2018}, some of which are: iterative refinement \citep{lee-emnlp-2018}, insertion-based methods \citep{stern-icml-2019,chan-neurips-2019,li-wngt-2019}, and conditional masked language models \citep{ghazvininejad-emnlp-2019,ghazvininejad-arxiv-2020}. Like insertion-based models \citep{stern-icml-2019,chan-arxiv-2019b}, our work does not commit to a fixed target length; insertion-based models can dynamically grow the canvas size, whereas our work which relies on a latent alignment can only generate a target sequence up to a fixed maximum predetermined length. Compared to conditional masked languages models \citep{ghazvininejad-emnlp-2019,ghazvininejad-arxiv-2020}, key differences are: 1) our models do not require target length prediction, and 2) we eschew the encoder-decoder neural architecture formulation, but rather rely on the single simple decoder architecture. 
KERMIT \citep{chan-arxiv-2019,chan-neurips-2019} also has a similar neural architecture as us; they also eschew the conventional encoder-decoder architecture and have a unified architecture. Our work relies on the superpositioning of the input and output sequences via the latent alignment, whereas KERMIT relies on concatenation to process the input and output sequences. Their work is also more focused on generative $p(x, y)$ modelling, whereas our work is focused on conditional modelling $p(y|x)$.

Our CTC work is closely related to and inspired heavily by \citet{libovicky-emnlp-2018}, which applied CTC single step generation models. The key difference is that our work used data distillation for training, and we find that distillation provides a significant boost in performance for our CTC models.

Finally, our work is closely related to the concurrent work of \citet{ghazvininejad-axe-cmlm} on AXE CMLM. Similar to our work, they also assume a latent alignment and use dynamic programming for learning. Their work focused on the single-step generation and demonstrated strong results, while we apply our models to both single step and iterative generation.

\begin{table*}[t]
\caption{\small BLEU comparison for various autoregressive and non-autoregressive models. Imputer is able to match the autoregressive Transformer baseline with just 4 generation steps. Numbers reported for Imputer trained with data distilled from big autoregressive transformer for En $\leftrightarrow$ De, and base transformer for En $\leftrightarrow$ Ro. }
\centering
\label{table:iterative_decoding}
\small
\begin{tabular}{lcccccc}
\toprule
& &   \multicolumn{2}{c}{\textbf{WMT'14}} & \multicolumn{2}{c}{\textbf{WMT'16}} \\
\bfseries Method & \bfseries Iterations \bfseries & \bfseries En$\rightarrow$De & \bfseries De$\rightarrow$En & \bfseries En$\rightarrow$Ro & \bfseries Ro$\rightarrow$En  \\
\midrule
\textit{Autoregressive} \\
% \quad GNMT \citep{wu-arxiv-2016} & $n$ & 24.6 & - \\
\quad Base Transformer & $n$ & 27.8 & 31.2 & 34.3 & 34.0  \\ % & 35.8 \\
\midrule
\textit{Non-Autoregressive} \\
\quad Insertion Transformer \citep{stern-icml-2019} & $\approx \log_2 n$ & 27.4 & - & - & - \\
\quad KERMIT \citep{chan-arxiv-2019} & $\approx \log_2 n$ & 27.8 & 30.7 & - & -\\
\quad Iterative Refinement \citep{lee-emnlp-2018} & 10 & 21.6 & 25.5 & 29.3 & 30.2\\
\quad Mask-Predict \citep{ghazvininejad-emnlp-2019} & 4 & 25.9 & 29.9 & 32.5 & 33.2 \\
 & 10 & 27.0 & 30.5 & 33.1 & 33.3\\
\quad SMART \citep{ghazvininejad-arxiv-2020} & 4 & 27.0 & 30.9 & - & - \\
 & 10 & 27.7 & 31.3 & - & -\\
\quad DisCo \citep{kasai2020parallel} & $\approx 4$ & 27.3 & 31.3 & 33.2 & 33.3 \\
\quad JM-NAT \citep{guo-etal-2020-jointly} & $4$ & 27.1 & 31.5 &  33.0 &  33.2 \\
\midrule
\textit{Our Work} \\
\quad Imputer & 2  & 27.5 & 30.6 & 33.7 & 33.4 \\
 & 4  & 28.0 & 31.5 & 34.3 & 34.0 \\
 & 8  & \textbf{28.2} & \textbf{31.8} & \textbf{34.4} & \textbf{34.1} \\
\bottomrule
\end{tabular}
 
\end{table*}

\section{Experiments}

\textbf{Hyperparameters.} We follow the base Transformer \cite{vaswani-nips-2017} for our experiments. However, since our architecture does not contain an encoder, we double the number of layers in our decoder to maintain the same number of parameters. Our models consist of 12 self-attention layers, with 512 hidden size, 2048 filter size, and 8 attention heads per layer. We use 0.1 dropout for regularization. We batch sequences of approximately same lengths together, with approximately 2048 tokens per batch. We use Adam optimizer \citep{kingma-iclr-2015} with $\beta = (0.9, 0.997)$ and $\epsilon = 10^{-9}$. The learning rate warms up to $10^{-3}$ in the first 10k steps and then decays with the inverse square root schedule following the Tensor2Tensor implementation \citep{tensor2tensor}. We train all our models for 2M steps. We train the Imputer using CTC loss (all masked prior alignment) for 1M steps, followed by Bernoulli masking policy \citep{chan2020imputer} for next 1M steps. We average the 5 checkpoints with the best performance on the development set to get the final model. For Imputer, we use top-\textit{k} decoding during inference.  We use canvas scale $s = 2$ for all our experiments, meaning we upsample the source sequence by a factor of 2.

\textbf{Dataset.} We perform experiments on WMT'14 En$\leftrightarrow$De, using newstest2013 as the development set, and report newstest2014 as the test set. We also report our performance on WMT'16 En-Ro. We use SentencePiece \citep{kudo-emnlp-2018} to generate a shared subword vocabulary. We evaluate the performance of our models with BLEU \citep{papineni2002bleu}.

\textbf{Distillation.} We follow prior work \citep{gu-iclr-2018, lee-emnlp-2018,stern-icml-2019,ghazvininejad-emnlp-2019} and use data distilled from an autoregressive teacher for training our models. We use autoregressive base Transformers for generating distilled data. For iterative generation, we also report the performance of Imputer model trained on data distilled from autoregressive big Transformers to be comparable with \citep{ghazvininejad-emnlp-2019,ghazvininejad-arxiv-2020} which distilled from a big Transformer. For WMT'16 En-Ro, we use the distilled dataset provided by \citet{ghazvininejad-emnlp-2019}\footnote{\scriptsize \url{https://github.com/facebookresearch/Mask-Predict}}. We analyze the impact of distillation on the performance of our models in Section \ref{sec:impactofdistillation}.

\subsection{Single Step Decoding}
We first report the performance of latent alignment models for single-step decoding. CTC makes full conditional independence assumption allowing the generation of the entire target sequence in a single step. We can also perform non-autoregressive single step generation with Imputer by imputing all of the tokens at once. Table \ref{table:1_step_decoding} summarizes the performance of our models and other non-autoregressive single step generation models. Our CTC model achieves 25.7 BLEU, and the Imputer model achieves 25.8 BLEU for WMT'14 En$\rightarrow$De. We find that our single step generation models outperform the autoregressive GNMT model of \citet{wu-arxiv-2016} on En$\rightarrow$De with 24.6 BLEU. To the best of our knowledge, our CTC and Imputer models outperform all prior work on single-step generation on WMT'14 En$\leftrightarrow$De and WMT'16 En$\leftrightarrow$Ro.

\subsection{Iterative Decoding}
We now analyze the performance of Imputer.
Imputer uses a constant number of decoding iterations independent of sequence length. We compare our performance with other sub-linear non-autoregressive models, ranging from models requiring logarithmic to a constant number of decoding iterations.
Table \ref{table:iterative_decoding} summarizes the results of Imputer model.

Our Imputer model using 8 decoding iterations achieves 28.2 BLEU on En$\rightarrow$De, slightly outperforming the autoregressive Transformer of 27.8 BLEU. On De$\rightarrow$En, we achieve 31.3 BLEU, on par with the autoregressive Transformer model. Similarly, on En$\leftrightarrow$Ro, Imputer matches the performance of the autoregressive teacher using just 4 decoding iterations. We also observe the robustness of our Imputer model when we reduce the number of decoding iterations from 8 to 2. Using only 2 iterations, we obtain 27.5 BLEU on En$\rightarrow$De and 30.2 BLEU on De$\rightarrow$En. These results were trained with distillation from a big Transformer model.
However, even when we distill from the base Transformer as shown in Table \ref{table:ar_decoding}, Imputer still performs on par with the autoregressive Transformer achieving 27.9 and 31.1 BLEU on En$\rightarrow$De and De$\rightarrow$En respectively. Figure \ref{fig:decoding} shows an example 8-step iterative decoding by Imputer.

\begin{table}[!t]
\centering
\caption{\small WMT'14 En-De BLEU comparison for distillation base vs big Transformer, and number of decoding iterations.}
\label{table:ar_decoding}
%\scalebox{0.7}{
\small
\begin{tabular}{@{\hspace{.1cm}}l@{\hspace{.15cm}}c@{\hspace{.15cm}}c@{\hspace{.15cm}}c@{\hspace{.1cm}}}
\toprule
\bfseries Model & Iterations &  \bfseries En$\rightarrow$De  &  \bfseries De$\rightarrow$En  \\
\midrule
Transformer \textit{(Base)} & $n$ & 27.8 & 31.2  \\
Transformer \textit{(Big)}  & $n$ & 29.5 & 32.2  \\
\midrule

Imputer \textit{(Base Distillation)} 
& 2 & 27.3 & 30.3 \\
& 4 & 27.9 & 30.9 \\
& 8 & 27.9 & 31.1 \\
& $n$ & 28.3 & 31.2 \\
\midrule
Imputer \textit{(Big Distillation)} 
& 2 & 27.5 & 30.2 \\
& 4 & 28.0 & 31.0 \\
& 8 & 28.2 &  31.3 \\
& $n$ & 28.4 &  31.4 \\
\bottomrule
\end{tabular}
%}
\end{table}

\begin{table}[t]
\centering
\caption{\small WMT'14 En$\leftrightarrow$De repeated token percentage comparison for single step generation models.}
\label{table:repeated_tokens}
\scalebox{0.7}{
\begin{tabular}{lccc}
\toprule
\bfseries Model  & \bfseries En$\rightarrow$De & \bfseries De$\rightarrow$En \\
\midrule
Gold Test Set & 0.04\% & 0.02\% \\
\midrule 
Mask-Predict \citep{ghazvininejad-emnlp-2019}  & 16.72\% & 12.31\%  \\
AXE CMLM \citep{ghazvininejad-axe-cmlm}  &  1.41\% & 1.03\%  \\
CTC (Our Work) &\textbf{ 0.17\%} & \textbf{0.23\%} \\
\bottomrule
\end{tabular}}
\end{table}

\section{Analysis}
%\st{In this section, we perform further analysis of our latent alignment models. We analyze the impact of the number of decoding iterations on Imputer, and the impact of distillation on our models.}
In this section, we present further analysis of our latent alignment models. We analyze the (1) impact on token repetitions in generated translations, (2) impact of the number of decoding iterations on Imputer, (3) impact of distillation on our models, and (4) impact of target length on Imputer.

\subsection{Token Repetitions}
We compare the repetition rate of our CTC model with single-step Mask-Predict \citep{ghazvininejad-emnlp-2019} and the concurrent work AXE CMLM \citep{ghazvininejad-axe-cmlm} in Table \ref{table:repeated_tokens}. We also report the percentage of repetitions in the original test set for reference. We observe a significantly lower rate of token repetitions in our CTC model compared to both the models. This empirical observation supports our hypothesis that $\beta^{-1}$ helps remove spurious token repetitions.

\begin{table}[t!]
\centering
\caption{\small Average relative decoding speed-up w.r.t. autoregressive greedy decoding baseline on WMT'14 En$\rightarrow$De test set for Imputer.}
\label{table:decoding_speed_up}
\scalebox{0.7}{
\begin{tabular}{cc}
\toprule
\bfseries Iterations  & \bfseries Relative Speed-Up \\
\midrule
1 (CTC / Imputer) & $\times 18.6$ \\
2 & $\times 9.2$ \\
4 & $\times 5.9$ \\
8 & $\times 3.9$ \\
\bottomrule
\end{tabular}}
\end{table}

\subsection{Impact of Number of Decoding Iterations}
The number of decoding iterations is an important hyperparameter in iterative models, providing a tunable trade-off between quality and inference speed. The ideal parallel decoding model should be robust to the number of decoding iterations, i.e. reducing the number of iterations should have minimal impact on performance. To analyze this capability of our Imputer model, we study the impact of the number of decoding iterations vs BLEU. We use the Imputer trained with data distilled from the autoregressive base Transformer teacher for this analysis. Imputer controls the number of decoding iterations through the top-$k$ hyperparameter, which imputes $k$ tokens per step. On one end, imputing all the tokens ($k=\infty$) in one step results in single-step decoding, while on the other end, imputing 1 token per step ($k=1$) results in linear autoregressive decoding (but not necessarily left-to-right).

\begin{figure}[!t]
\vspace{-.2cm}
    \centering
    \resizebox{.66\linewidth}{!}{{\pgfplotsset{compat=1.3}
\begin{tikzpicture}
\begin{axis}[
    title={BLEU vs. Number of Decoding Iterations},
    xlabel={\# Decoding Iterations},
    ylabel={BLEU},
    ymin=24.8, ymax=29.0,
    xmin=0, xmax=8,
    xticklabels={1,2,4,8,16,32,\textit{N}}, xtick={1,2,3,4,5,6,7},
    ytick={25,26,27,28,29},
    ymajorgrids=true,
    grid style=dashed,
    legend pos=south east,
    legend cell align=left,
    every axis plot/.append style={thick}
]
\addplot[
    color=blue,
    mark=*,
    ]
    coordinates {
    (1,25.8)(2,27.3)(3,27.85)(4,27.89)(5,28.1)(6,28.2)(7,28.3)
    };
\addplot[color=red, dashed] coordinates {(0,27.8) (8,27.8)};
\legend{Imputer, Autoregressive Teacher}
\end{axis}
\end{tikzpicture}}}
    \caption{\small WMT'14 En$\rightarrow$De BLEU comparison for different number of decoding iterations for Imputer.}
    \label{fig:decode_iter_vs_bleu}
%\vspace{-.2cm}
\end{figure}
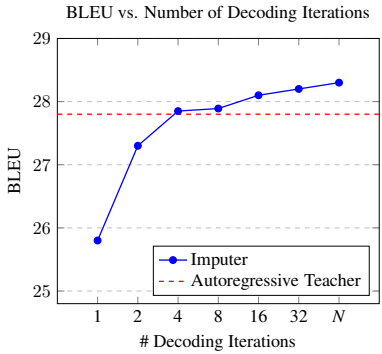

Figure \ref{fig:decode_iter_vs_bleu} shows the BLEU score vs target length $T$ for WMT'14 En$\rightarrow$De test set, where $T$ is the number of decoding iterations. As expected, the performance consistently increases with an increase in $T$. We find that Imputer is robust to $T$, sacrificing just 0.6 BLEU points when reducing $T$ from 8 to 2. We can match the performance of its autoregressive teacher using just 4 decoding iterations. Interestingly, the performance keeps increasing consistently beyond 8 iterations, and even outperforming the autoregressive teacher %itself
slightly. In the extreme case of autoregressive $O(n)$ decoding, we obtain 28.3 BLEU score, exceeding the teacher's performance by 0.5 BLEU points.

\subsection{Impact of Distillation}
\label{sec:impactofdistillation}

\begin{table}[!t]
\centering
\caption{\small WMT'14 En$\rightarrow$De BLEU comparison showing the impact of distillation.}
\label{table:raw_vs_distilled}
\small
\begin{tabular}{lcccc}
\toprule
\bfseries Method & \bfseries Iterations & \bfseries Original &  \bfseries Distillation \\
\midrule
CTC & 1 & 15.6 & 25.4 \\
Imputer & 4 & 24.7 & 27.9 \\
& 8 & 25.0 & 27.9 \\
\bottomrule
\end{tabular}
\vspace{-0.2cm}
\end{table}

We analyze the impact of distillation on our models by comparing them to original training data versus training data from a base Transformer teacher on the WMT'14 En$\rightarrow$De dataset. Table \ref{table:raw_vs_distilled} summarizes the results. In all cases, models trained with the distilled data perform significantly better than the model trained with the original data. We observe that the performance gap is largest in the case of the CTC model, and decreases with an increase in the number of decoding iterations. This is consistent with prior work finding distillation to improve model quality \citep{gu-iclr-2018,zhou2019understanding}.

\subsection{Impact of Target Length for Imputer}

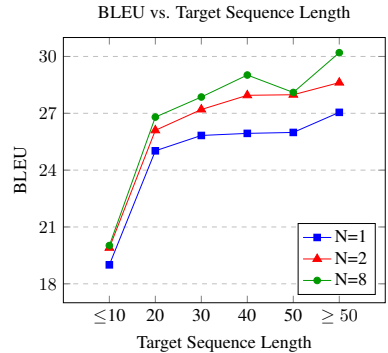
\begin{figure}[h]
    \centering
    \resizebox{.66\linewidth}{!}{{\pgfplotsset{compat=1.3}
\begin{tikzpicture}
\begin{axis}[
    title={BLEU vs. Target Sequence Length},
    xlabel={Target Sequence Length},
    ylabel={BLEU},
    ymin=17.0, ymax=31.0,
    xmin=0, xmax=7,
    xticklabels={$\leq$10,20,30,40,50,$\geq50$}, xtick={1,2,3,4,5,6},
    ytick={18,21,24,27,30},
    ymajorgrids=true,
    grid style=dashed,
    legend pos=south east,
    legend cell align=left,
    every axis plot/.append style={ultra thin}
]
\addplot[ mark=square*, mark size=2,
    color=blue,
    ]
    coordinates {
    (1,19.0)(2,25.02)(3,25.83)(4,25.94)(5,25.99)(6,27.05)
    };
\addplot[ mark=triangle*, mark size=3,
    color=red,
    ]
    coordinates {
    (1,19.9)(2,26.1)(3,27.2)(4,27.95)(5,27.98)(6,28.62)
    };
\addplot[mark=*, mark size=2,
    color=green!60!black,
    ]
    coordinates {
    (1,20.02)(2,26.8)(3,27.86)(4,29.02)(5,28.1)(6,30.20)
    };
    \legend{N=1, N=2, N=8}
\end{axis}
\end{tikzpicture}}}
    \caption{\small WMT'14 En$\rightarrow$De BLEU comparison for sentences binned by target sequence length for Imputer; $N$ is the number of decoding iterations.}
    \label{fig:sent_length_vs_bleu}
\vspace{-.2cm}
\end{figure}

Figure \ref{fig:sent_length_vs_bleu} depicts the impact of number of decoding iterations bucketed by the target sequence length $N$. We use the \texttt{compare-mt} \cite{compare_mt} package to bucket test set examples based on target sentence length, and compute BLEU score for each bucket using a different number of decoding iterations. Increase in the number of decoding iterations provides consistent gain across all buckets.

%\vspace{0.5cm}

\section{Conclusion}
In this paper, we investigated two latent alignments models, CTC and Imputer, for non-autoregressive machine translation. CTC is a single step generation model, while Imputer is an iterative generative model requiring only a constant number of generation steps. Our models rely on dynamic programming to marginalize out the latent alignments. Unlike many prior works, our models do not need to perform target length prediction, or re-scoring of candidates and our models use a simplified neural architecture without the need of cross-attention mechanism found in many prior encoder-decoder architectures. 
We demonstrate the ease and effectiveness of the application of these simple latent alignment models primarily used in speech recognition to the task of machine translation. Applying these latent alignment models for parallel translation of long documents can be an interesting research direction. 

\section*{Acknowledgments}
We give thanks to Colin Cherry, Geoffrey Hinton, George Foster, Jakob Uszkoreit, Jamie Kiros, Julia Kreutzer, Samy Bengio, and Sara Sabour for research discussions and technical assistance.

\bibliography{paper}

\begin{thebibliography}{36}
\expandafter\ifx\csname natexlab\endcsname\relax\def\natexlab#1{#1}\fi

\bibitem[{Chan et~al.(2019{\natexlab{a}})Chan, Kiros, and
  Chan}]{chan-neurips-2019}
Harris Chan, Jamie Kiros, and William Chan. 2019{\natexlab{a}}.
\newblock {Multilingual KERMIT: It’s Not Easy Being Generative}.
\newblock In \emph{{NeurIPS: Workshop on Perception as Generative Reasoning}}.

\bibitem[{Chan et~al.(2019{\natexlab{b}})Chan, Kitaev, Guu, Stern, and
  Uszkoreit}]{chan-arxiv-2019}
William Chan, Nikita Kitaev, Kelvin Guu, Mitchell Stern, and Jakob Uszkoreit.
  2019{\natexlab{b}}.
\newblock {KERMIT: Generative Insertion-Based Modeling for Sequences}.
\newblock In \emph{{arXiv}}.

\bibitem[{Chan et~al.(2020)Chan, Saharia, Hinton, Norouzi, and
  Jaitly}]{chan2020imputer}
William Chan, Chitwan Saharia, Geoffrey Hinton, Mohammad Norouzi, and Navdeep
  Jaitly. 2020.
\newblock Imputer: Sequence modelling via imputation and dynamic programming.
\newblock In \emph{{arXiv}}.

\bibitem[{Chan et~al.(2019{\natexlab{c}})Chan, Stern, Kiros, and
  Uszkoreit}]{chan-arxiv-2019b}
William Chan, Mitchell Stern, Jamie Kiros, and Jakob Uszkoreit.
  2019{\natexlab{c}}.
\newblock {An Empirical Study of Generation Order for Machine Translation}.
\newblock In \emph{{arXiv}}.

\bibitem[{Dyer et~al.(2013)Dyer, Chahuneau, and Smith}]{dyer-naacl-2013}
Chris Dyer, Victor Chahuneau, and Noah Smith. 2013.
\newblock {A Simple, Fast, and Effective Reparameterization of IBM Model 2}.
\newblock In \emph{{NAACL}}.

\bibitem[{Ghazvininejad et~al.(2020{\natexlab{a}})Ghazvininejad, Karpukhin,
  Zettlemoyer, and Levy}]{ghazvininejad-axe-cmlm}
Marjan Ghazvininejad, Vladimir Karpukhin, Luke Zettlemoyer, and Omer Levy.
  2020{\natexlab{a}}.
\newblock {Aligned Cross Entropy for Non-Autoregressive Machine Translation}.
\newblock In \emph{{arXiv}}.

\bibitem[{Ghazvininejad et~al.(2019)Ghazvininejad, Levy, Liu, and
  Zettlemoyer}]{ghazvininejad-emnlp-2019}
Marjan Ghazvininejad, Omer Levy, Yinhan Liu, and Luke Zettlemoyer. 2019.
\newblock {Mask-Predict: Parallel Decoding of Conditional Masked Language
  Models}.
\newblock In \emph{{EMNLP}}.

\bibitem[{Ghazvininejad et~al.(2020{\natexlab{b}})Ghazvininejad, Levy, and
  Zettlemoyer}]{ghazvininejad-arxiv-2020}
Marjan Ghazvininejad, Omer Levy, and Luke Zettlemoyer. 2020{\natexlab{b}}.
\newblock {Semi-Autoregressive Training Improves Mask-Predict Decoding}.
\newblock In \emph{{arXiv}}.

\bibitem[{Graves et~al.(2006)Graves, Fernandez, Gomez, and
  Schmidhuber}]{graves-icml-2006}
Alex Graves, Santiago Fernandez, Faustino Gomez, and Jurgen Schmidhuber. 2006.
\newblock {Connectionist Temporal Classification: Labelling Unsegmented
  Sequence Data with Recurrent Neural Networks}.
\newblock In \emph{{ICML}}.

\bibitem[{Graves and Jaitly(2014)}]{graves-icml-2014}
Alex Graves and Navdeep Jaitly. 2014.
\newblock {Towards End-to-End Speech Recognition with Recurrent Neural
  Networks}.
\newblock In \emph{{ICML}}.

\bibitem[{Graves et~al.(2013)Graves, Mohamed, and Hinton}]{graves-icassp-2013}
Alex Graves, {Abdel-rahman} Mohamed, and Geoffrey Hinton. 2013.
\newblock {Speech Recognition with Deep Recurrent Neural Networks}.
\newblock In \emph{{ICASSP}}.

\bibitem[{Gu et~al.(2018)Gu, Bradbury, Xiong, Li, and Socher}]{gu-iclr-2018}
Jiatao Gu, James Bradbury, Caiming Xiong, Victor~O.K. Li, and Richard Socher.
  2018.
\newblock {Non-Autoregressive Neural Machine Translation}.
\newblock In \emph{{ICLR}}.

\bibitem[{Gu et~al.(2019)Gu, Wang, and Zhao}]{gu-neurips-2019}
Jiatao Gu, Changhan Wang, and Jake Zhao. 2019.
\newblock {Levenshtein Transformer}.
\newblock In \emph{{NeurIPS}}.

\bibitem[{Guo et~al.(2020)Guo, Xu, and Chen}]{guo-etal-2020-jointly}
Junliang Guo, Linli Xu, and Enhong Chen. 2020.
\newblock \href {https://doi.org/10.18653/v1/2020.acl-main.36} {Jointly masked
  sequence-to-sequence model for non-autoregressive neural machine
  translation}.
\newblock In \emph{Proceedings of the 58th Annual Meeting of the Association
  for Computational Linguistics}, pages 376--385, Online. Association for
  Computational Linguistics.

\bibitem[{Kasai et~al.(2020)Kasai, Cross, Ghazvininejad, and
  Gu}]{kasai2020parallel}
Jungo Kasai, James Cross, Marjan Ghazvininejad, and Jiatao Gu. 2020.
\newblock {Parallel Machine Translation with Disentangled Context Transformer}.
\newblock \emph{{arXiv preprint arXiv:2001.05136}}.

\bibitem[{Kingma and Ba(2015)}]{kingma-iclr-2015}
Diederik Kingma and Jimmy Ba. 2015.
\newblock {Adam: A Method for Stochastic Optimization}.
\newblock In \emph{{ICLR}}.

\bibitem[{Kudo and Richardson(2018)}]{kudo-emnlp-2018}
Taku Kudo and John Richardson. 2018.
\newblock {SentencePiece: A simple and language independent subword tokenizer
  and detokenizer for Neural Text Processing}.
\newblock In \emph{{EMNLP}}.

\bibitem[{Lee et~al.(2018)Lee, Mansimov, and Cho}]{lee-emnlp-2018}
Jason Lee, Elman Mansimov, and Kyunghyun Cho. 2018.
\newblock {Deterministic Non-Autoregressive Neural Sequence Modeling by
  Iterative Refinement}.
\newblock In \emph{{EMNLP}}.

\bibitem[{Li and Chan(2019)}]{li-wngt-2019}
Lala Li and William Chan. 2019.
\newblock {Big Bidirectional Insertion Representations for Documents}.
\newblock In \emph{{EMNLP: Workshop of Neural Generation and Translation}}.

\bibitem[{Li et~al.(2019)Li, Lin, He, Tian, Qin, Wang, and Liu}]{li2019hint}
Zhuohan Li, Zi~Lin, Di~He, Fei Tian, Tao Qin, Liwei Wang, and Tie-Yan Liu.
  2019.
\newblock {Hint-Based Training for Non-Autoregressive Machine Translation }.
\newblock In \emph{{EMNLP}}.

\bibitem[{Libovicky and Helcl(2018)}]{libovicky-emnlp-2018}
Jindrich Libovicky and Jindrich Helcl. 2018.
\newblock {End-to-End Non-Autoregressive Neural Machine Translation with
  Connectionist Temporal Classification}.
\newblock In \emph{{EMNLP}}.

\bibitem[{Liu et~al.(2020)Liu, Ren, Tan, Zhang, Qin, Zhao, and
  Liu}]{liu2020task}
Jinglin Liu, Yi~Ren, Xu~Tan, Chen Zhang, Tao Qin, Zhou Zhao, and Tie-Yan Liu.
  2020.
\newblock Task-level curriculum learning for non-autoregressive neural machine
  translation.
\newblock \emph{arXiv preprint arXiv:2007.08772}.

\bibitem[{Ma et~al.(2019)Ma, Zhou, Li, Neubig, and Hovy}]{ma2019flowseq}
Xuezhe Ma, Chunting Zhou, Xian Li, Graham Neubig, and Eduard Hovy. 2019.
\newblock {FlowSeq: Non-Autoregressive Conditional Sequence Generation with
  Generative Flow}.
\newblock In \emph{{EMNLP}}.

\bibitem[{Manning et~al.(1999)Manning, Manning, and
  Sch{\"u}tze}]{manning1999foundations}
Christopher~D Manning, Christopher~D Manning, and Hinrich Sch{\"u}tze. 1999.
\newblock \emph{Foundations of statistical natural language processing}.
\newblock MIT press.

\bibitem[{Neubig et~al.(2019)Neubig, Dou, Hu, Michel, Pruthi, Wang, and
  Wieting}]{compare_mt}
Graham Neubig, Zi{-}Yi Dou, Junjie Hu, Paul Michel, Danish Pruthi, Xinyi Wang,
  and John Wieting. 2019.
\newblock {compare-mt: {A} Tool for Holistic Comparison of Language Generation
  Systems}.
\newblock \emph{{CoRR}}.

\bibitem[{Papineni et~al.(2002)Papineni, Roukos, Ward, and
  Zhu}]{papineni2002bleu}
Kishore Papineni, Salim Roukos, Todd Ward, and Wei-Jing Zhu. 2002.
\newblock {BLEU: a Method for Automatic Evaluation of Machine Translation}.
\newblock In \emph{{ACL}}.

\bibitem[{Ruis et~al.(2019)Ruis, Stern, Proskurnia, and Chan}]{ruis-wngt-2019}
Laura Ruis, Mitchell Stern, Julia Proskurnia, and William Chan. 2019.
\newblock {Insertion-Deletion Transformer}.
\newblock In \emph{{EMNLP: Workshop of Neural Generation and Translation}}.

\bibitem[{Shao et~al.(2020)Shao, Zhang, Feng, Meng, and
  Zhou}]{shao2019minimizing}
Chenze Shao, Jinchao Zhang, Yang Feng, Fandong Meng, and Jie Zhou. 2020.
\newblock {Minimizing the Bag-of-Ngrams Difference for Non-Autoregressive
  Neural Machine Translation}.
\newblock In \emph{{AAAI}}.

\bibitem[{Stern et~al.(2019)Stern, Chan, Kiros, and
  Uszkoreit}]{stern-icml-2019}
Mitchell Stern, William Chan, Jamie Kiros, and Jakob Uszkoreit. 2019.
\newblock {Insertion Transformer: Flexible Sequence Generation via Insertion
  Operations}.
\newblock In \emph{{ICML}}.

\bibitem[{Sun et~al.(2019)Sun, Li, Wang, He, Lin, and Deng}]{sun2019fast}
Zhiqing Sun, Zhuohan Li, Haoqing Wang, Di~He, Zi~Lin, and Zhihong Deng. 2019.
\newblock {Fast Structured Decoding for Sequence Models}.
\newblock In \emph{{NeurIPS}}.

\bibitem[{Sun and Yang(2020)}]{sun2020approach}
Zhiqing Sun and Yiming Yang. 2020.
\newblock {An EM Approach to Non-autoregressive Conditional Sequence
  Generation}.
\newblock \emph{{arXiv preprint arXiv:2006.16378}}.

\bibitem[{Vaswani et~al.(2018)Vaswani, Bengio, Brevdo, Chollet, Gomez, Gouws,
  Jones, Kaiser, Kalchbrenner, Parmar, Sepassi, Shazeer, and
  Uszkoreit}]{tensor2tensor}
Ashish Vaswani, Samy Bengio, Eugene Brevdo, Francois Chollet, Aidan~N. Gomez,
  Stephan Gouws, Llion Jones, \L{}ukasz Kaiser, Nal Kalchbrenner, Niki Parmar,
  Ryan Sepassi, Noam Shazeer, and Jakob Uszkoreit. 2018.
\newblock {Tensor2Tensor for Neural Machine Translation}.
\newblock In \emph{{AMTA}}.

\bibitem[{Vaswani et~al.(2017)Vaswani, Shazeer, Parmar, Uszkoreit, Jones,
  Gomez, Kaiser, and Polosukhin}]{vaswani-nips-2017}
Ashish Vaswani, Noam Shazeer, Niki Parmar, Jakob Uszkoreit, Llion Jones,
  Aidan~N. Gomez, Lukasz Kaiser, and Illia Polosukhin. 2017.
\newblock {Attention Is All You Need}.
\newblock In \emph{{NIPS}}.

\bibitem[{Wang et~al.(2019)Wang, Tian, He, Qin, Zhai, and Liu}]{wang2019non}
Yiren Wang, Fei Tian, Di~He, Tao Qin, ChengXiang Zhai, and Tie-Yan Liu. 2019.
\newblock {Non-Autoregressive Machine Translation with Auxiliary
  Regularization}.
\newblock In \emph{{AAAI}}.

\bibitem[{Wu et~al.(2016)Wu, Schuster, Chen, Le, Norouzi, Macherey, Krikun,
  Cao, Gao, Macherey, Klingner, Shah, Johnson, Liu, Łukasz Kaiser, Gouws,
  Kato, Kudo, Kazawa, Stevens, Kurian, Patil, Wang, Young, Smith, Riesa,
  Rudnick, Vinyals, Corrado, Hughes, and Dean}]{wu-arxiv-2016}
Yonghui Wu, Mike Schuster, Zhifeng Chen, Quoc~V. Le, Mohammad Norouzi, Wolfgang
  Macherey, Maxim Krikun, Yuan Cao, Qin Gao, Klaus Macherey, Jeff Klingner,
  Apurva Shah, Melvin Johnson, Xiaobing Liu, Łukasz Kaiser, Stephan Gouws,
  Yoshikiyo Kato, Taku Kudo, Hideto Kazawa, Keith Stevens, George Kurian,
  Nishant Patil, Wei Wang, Cliff Young, Jason Smith, Jason Riesa, Alex Rudnick,
  Oriol Vinyals, Greg Corrado, Macduff Hughes, and Jeffrey Dean. 2016.
\newblock {Google's Neural Machine Translation System: Bridging the Gap between
  Human and Machine Translation}.
\newblock In \emph{{arXiv}}.

\bibitem[{Zhou et~al.(2020)Zhou, Neubig, and Gu}]{zhou2019understanding}
Chunting Zhou, Graham Neubig, and Jiatao Gu. 2020.
\newblock {Understanding Knowledge Distillation in Non-autoregressive Machine
  Translation}.
\newblock In \emph{{ICLR}}.

\end{thebibliography}
\bibliographystyle{acl_natbib}

\appendix

\clearpage

\end{document}